\pdfoutput=1
\documentclass[11pt]{article}

\usepackage[final]{coling}

\usepackage{times}
\usepackage{latexsym}

\usepackage[T1]{fontenc}
\usepackage[utf8]{inputenc}

\usepackage{microtype}

\usepackage{inconsolata}

\usepackage{graphicx}

\usepackage{algorithm}
\usepackage{algorithmic}
\usepackage{multirow}
\usepackage{subfigure}
\usepackage{booktabs}
\usepackage{amsmath}

\title{Boosting Meta-Learning for Few-Shot Text Classification via Label-guided Distance Scaling}

\author{Yunlong Gao, Xinyue Liu\thanks{*Corresponding author}, Yingbo Wang, Linlin Zong, Bo Xu \\
        Dalian University of Technology, Dalian, China\\
         yvogao@mail.dlut.edu.cn, 
         xyliu@dlut.edu.cn, \\
         yingbo.wang.dlut.edu@outlook.com,
         llzong@dlut.edu.cn, xubo@dlut.edu.cn  \ }

\begin{document}
\maketitle
\begin{abstract}

Few-shot text classification aims to recognize unseen classes with limited labeled text samples. Existing approaches focus on boosting meta-learners by developing complex algorithms in the training stage. However, the labeled samples are randomly selected during the testing stage, so they may not provide effective supervision signals, leading to misclassification. 
To address this issue, we propose a \textbf{L}abel-guided \textbf{D}istance \textbf{S}caling (LDS) strategy.
The core of our method is exploiting label semantics as supervision signals in both the training and testing stages.
Specifically, in the training stage, we design a label-guided loss to inject label semantic information, pulling closer the sample representations and corresponding label representations.
In the testing stage, we propose a Label-guided Scaler which scales sample representations with label semantics to provide additional supervision signals.
Thus, even if labeled sample representations are far from class centers, our Label-guided Scaler pulls them closer to their class centers, thereby mitigating the misclassification.
We combine two common meta-learners to verify the effectiveness of the method. 
Extensive experimental results demonstrate that our approach significantly outperforms state-of-the-art models.
% All datasets and codes are available at https://anonymous.4open.science/r/Label-guided-Text-Classification.

\end{abstract}

\section{Introduction}

Text classification has been widely applied to various real applications, such as intent detection~\citep{Dopierre_Gravier_Logerais_2021}, news classification~\citep{news}, and so on. However, a deficiency of supervised data is often experienced in many real-world scenarios, thereby promoting the emergence of few-shot text classification~\citep{Yu_2018, Pan_2019}, which aims to detect unseen classes with limited labeled text samples by using knowledge learned from seen classes.

\begin{figure}

\includegraphics[width=1\linewidth]{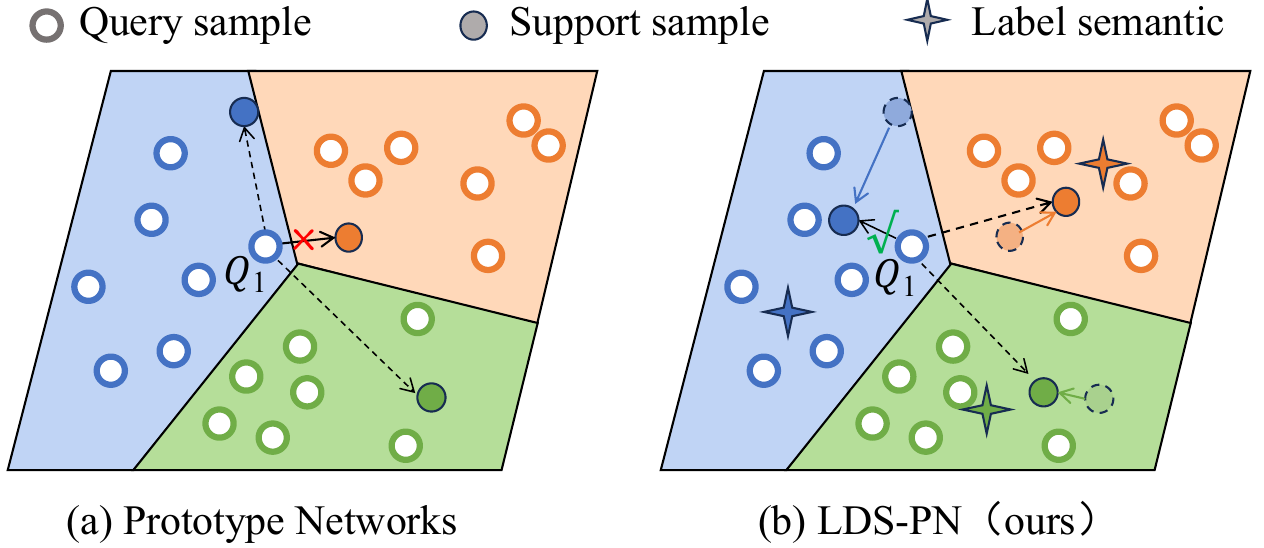}
\caption{Illustration of the problem in the testing stage, the feature space classification diagram of Prototypical Networks and the proposed LDS-PN.
Picture (a) shows that, due to the support sample of class \textit{blue} is at the boundary of class distribution, and $Q_1$ is misclassified as class \textit{orange} because it is closest to the support sample of class \textit{orange}. Picture (b) shows LDS-PN pulls support samples closer to the center of the corresponding classes by leveraging their label semantics (names), so $Q_1$ is classified class \textit{blue} correctly.}
\label{fig1}
\vspace{-0.3cm}
\end{figure}

% \begin{figure}[!t]
% \subfigure[Prototypical Networks]{
% % \subfigure[Traditional measurement methods]{
% \label{fig1a}
% \includegraphics[width=0.22\textwidth]{figure/pro1.pdf}}
% \subfigure[LDS-PN]{
% \label{fig1b}
% \includegraphics[width=0.22\textwidth]{figure/pro2.pdf}}
% \caption{The feature space classification diagram of Prototypical Networks and the proposed LDS-PN. There are three classes: \textit{BLUE}, \textit{ORANGE}, and \textit{GREEN}. The solid circles $S_b$, $S_o$, and $S_g$ are \textit{BLUE}, \textit{ORANGE}, and \textit{GREEN} class support samples, respectively. The hollow circle $Q_1$ is a query sample, it belongs to class \textit{BLUE} in reality. Picture (a) shows that, due to unsatisfactory representation, $S_b$ is misrepresented in the distribution of class \textit{ORANGE}, and $Q_1$ is misclassified as class \textit{ORANGE} because it is closer to $S_o$ than $S_b$. Picture (b) shows LDS-PN pulls $S_b$, $S_o$, and $S_g$ into the distribution space of the corresponding class with their label $L_b$, $L_o$, and $L_g$, so $Q_1$ is classified correctly.}
% \label{fig1}
% \end{figure}

Existing methods primarily follow the meta-learning paradigm and focus on learning sample representations to distinguish different classes, and widely combine Prototypical Networks (PN)~\citep{proto_2017}, facilitating the classification of query samples by measuring their distances to support samples
% \footnote{In the few-shot setting, labeled samples are called support samples, and unlabeled samples are called query samples.}
. DE~\citep{Liu2023} and MEDA~\citep{MEDA} leverage data augmentation to expand the number of support samples. ContrastNet~\citep{Chen_Zhang_Mao_Xue_2022} introduces a multistage contrastive loss to reduce the distance between samples of the same class and increase the distance between samples of different classes, thereby creating more distinguishable sample representations. TART \citep{TART} transforms the class prototypes to per-class fixed reference points in task-adaptive metric spaces, making class prototypes distinguishable.

%问题
% Although previous works have achieved certain improvements over PN, it is worth noting that they directly use sample representations obtained by the model after training. Due to the labeled samples are randomly selected in the testing phase, they may not provide effective supervision signals for label prediction. As shown in Fig.~\ref{fig1}(a), in the representation space, if the support sample of the class \textit{blue} is at the boundary of the distribution of class \textit{blue}, the query sample $Q_1$ will be misclassified as class \textit{orange} because it is closest to the support sample of class \textit{orange}. As a result, this randomness may cause misclassification in the testing stage.
Although previous works have achieved certain improvements over PN, there is still a limitation in the testing stage. Specifically, the support samples are randomly selected in the testing phase, so they may not provide effective supervision signals for label prediction. As shown in Fig.~\ref{fig1}(a), in the representation space, if the support sample of the class \textit{blue} is at the boundary of the distribution of class \textit{blue}, the query sample $Q_1$ will be misclassified as class \textit{orange} because it is closest to the support sample of class \textit{orange}. It is worth noting that, existing methods only concentrate on designing powerful models at the training stage, but they neglect this problem in the testing stage. Even though they obtain high-quality representations, they may encounter misclassification caused by the randomly selected support samples in the testing stage.

Tackling this misclassification problem requires additional information (e.g., label semantics) in the testing stage. 
To achieve this, the additional information and samples need to have a correlation, and a guider is needed to exploit the additional information for the classification. % should太绝对
However, this solution may face some challenges. 
Firstly, due to the setting of meta-learning, the guider only can influence support samples in the training stage.
Hence, the additional information also needs to be injected through a specific training objective, influencing both support and query samples.
Secondly, in the testing stage, a trained guider only can project the support sample representations into another representation space, causing a different distribution for support samples and query samples.
Besides, due to the scarcity of support samples, training a guider will trap in overfitting.
Consequently, the guider should be a non-parametric method.

% Secondly, for a specific class, labeled samples and unlabeled samples should be in the same distribution.
% Due to the setting of meta-learning, if the additional information is injected by inputting the model, it only influences labeled samples, but not unlabeled samples. 
% Thus, this causes a different distribution for labeled samples and unlabeled samples. 
% Consequently, the additional information should not be injected by inputting the model, but should be in other forms (e.g., training objective) to consider both labeled and unlabeled samples.

% 第二个testing阶段，只能用EM，不能用训练的
% If additional information and labeled samples , the labeled samples will be closer to the additional information after training, and thus deviating the original distribution.
% However, since unlabeled samples cannot be guided, they still lie in the original distribution.
% Thus, a trained guide causes a different distributions for labeled samples and unlabeled samples.
% Consequently, a trained guide induces a negative influence for the classification.

We propose to leverage label semantics to tackle the above problem and present a \textbf{L}abel-guided \textbf{D}istance \textbf{S}caling (LDS) strategy.
We first utilize prompt learning to establish the correlation between samples and label semantics.
% To strengthen the correlation, 
Then, we design a label-guided loss to inject the label semantics in the training stage, which pulls closer to the sample representations and corresponding label representations. 
% Seeking for a non-parametric guider to exploit the label semantic information for the classification, we propose a Label-guided Scaler in the testing stage, which mixes support sample representations and label semantic information by EM algorithm, pulling support sample representations further closer to the class center.
After injecting label semantics into samples and establishing their correlation, we develop a guider in the testing stage to fully exploit the label semantics for the classification.
Specifically, we propose a Label-guided Scaler, which scales support sample representations with label semantics, pulling support sample representations further closer to the class centers.
Thus, even if labeled samples are far from class centers, our Label-guided Scaler pulls them closer to their class centers, thereby mitigating the misclassification, as shown in Fig.~\ref{fig1}(b).
% , the support samples are drawn closer to the class center, and the query sample $Q_1$ is closest to the support sample of class \textit{blue}, so it is classified correctly. 

%解决方案

To summarize, our work makes the following contributions:
(1) We point out the misclassification caused by the randomly selected support samples and present it should be solved in the testing stage with the additional information.
(2) We propose a \textbf{L}abel-guided \textbf{D}istance \textbf{S}caling (LDS) strategy including a label-guided loss and a Label-guided Scaler, making the class distributions more distinguishable.
(3) We conduct extensive experiments and results show that our approach significantly outperforms state-of-the-art models. 
Particularly, it achieves an average improvements of 9.4\% and 10.1\% in 5-way 1-shot and 10\&15-way 1-shot tasks.
(4) Our proposed LDS strategy not only boosts the performance of PN (metric-based method) but also could be used to boost other meta-learners, like RRML~\citep{BertinettoHTV19} for few-shot text classification tasks.

\section{Related Work}
\subsection{Meta-learning Based Methods}

Meta-learning has achieved remarkable success in few-shot learning~\citep{Hou_Chang_Ma_Shan_Chen_2019,match_2016,Liu_Fu_Xu_Yang_Li_Wang_Zhang_2022}, and it can be divided into two categories: (1) \textbf{Optimization-based methods}: These methods aim to find an optimal initialization parameter that can adapt to few-shot training examples, such as MAML~\citep{MAML_2017} and Reptile~\citep{MAML++_2018}. (2) \textbf{Metric-based methods}: These methods focus on learning a metric to measure similarities between samples and classes, including Prototypical Networks~\citep{proto_2017}, Relation Networks~\citep{relation_2017} and so on. Existing methods mainly fall into the Metric-based method, and focus on getting the distinguishable class distributions through training, like ContrastNet~\citep{Chen_Zhang_Mao_Xue_2022}, SPCNet~\citep{SPContrastNet}, or learning expanding the support samples, like MEDA~\citep{MEDA}, DE~\citep{Liu2023} and RPOTAUG~\citep{Dopierre_Gravier_Logerais_2021}. The key idea of our approach is leveraging label semantics to guide sample representations closer to the class centers in the testing stage, thereby reducing prediction inconsistencies caused by randomly selected support samples.

\subsection{Transfer Learning Based Methods}
Transfer learning aims to tackle few-shot text classification by leveraging knowledge from source domains, including two categories: (1) \textbf{Fine-tuning methods}: These methods aim to fine-tune Pre-trained Language Models (PLMs)~\citep{Devlin_2019, ROBERT_2019} and apply it to downstream tasks. (2) \textbf{Prompt learning methods}: These methods~\citep{gpt_2020, KPT} convert the classification task to a cloze-style mask language modeling problem and have recently gained attention. Several studies have applied prompt learning to various tasks, including knowledge probing~\citep{Feldman2019CommonsenseKM,Petroni_Rocktäschel_Riedel_Lewis_Bakhtin_Wu_Miller_2019}, text classification~\citep{Schick_2021,KPT++}, relation extraction~\citep{Han2021PTRPT,HeHMGLC23}, and entity typing~\citep{Han_Zhang_Ding_Gu_Liu_Huo_Qiu_Yao_Zhang_Zhang_2021,DBLP:conf/acl/0001TWZZX0Z23}.
These methods seek to design a prompt template or expand the label words to improve the ability of large-scale Language Models. By contrast, our work leverages prompt learning merely to strengthen the correlation between samples and the label semantics.

\begin{figure*}[h]
\centering
\includegraphics[width=.9\textwidth]{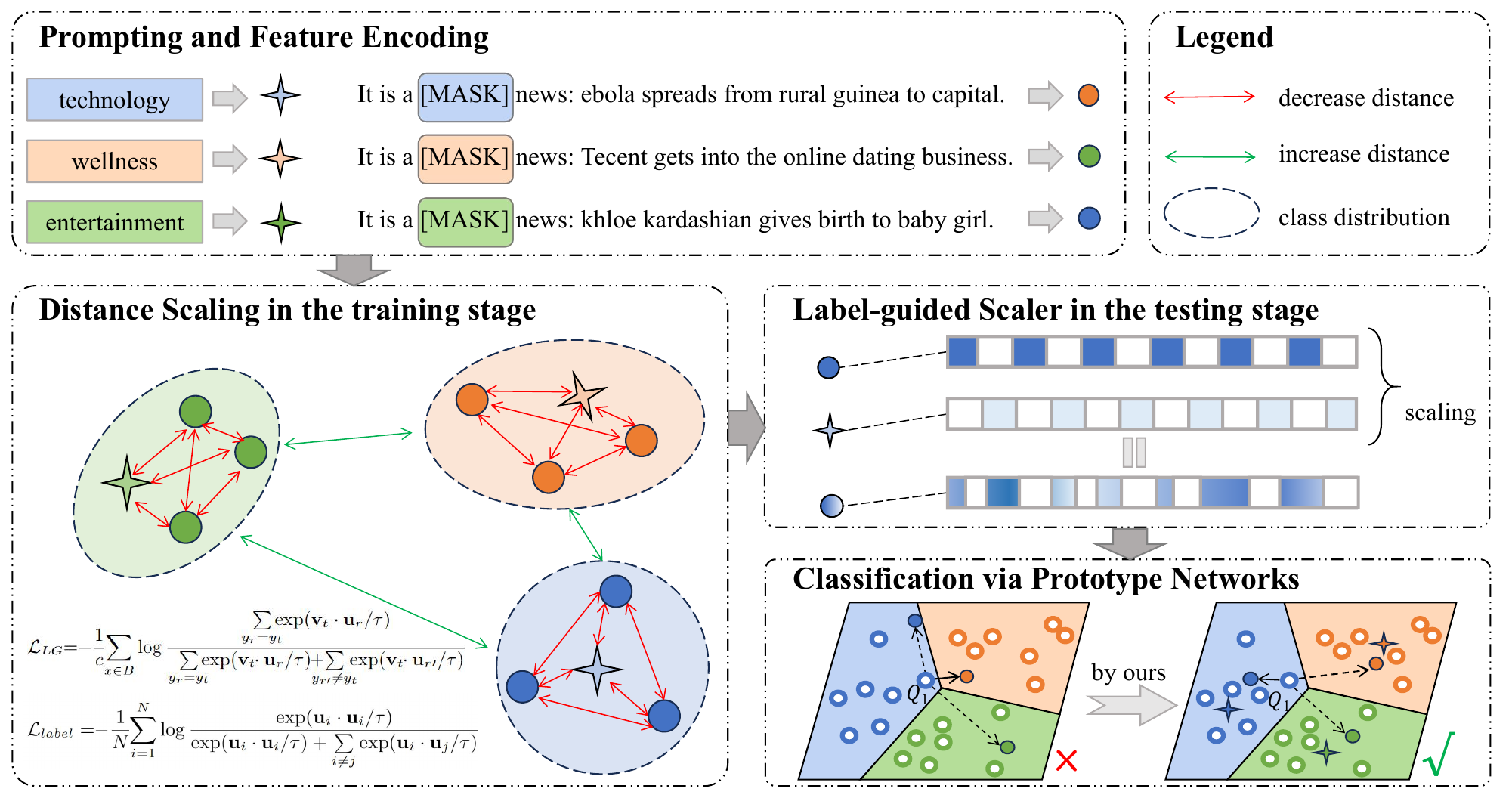} % Reduce the figure size so that it is slightly narrower than the column.
\caption{The graph illustrates the Prompting and Feature Encoding method, Distance Scaling in the training stage, Label-guide Scaler in the testing stage, and Classification via Prototype Networks.}
\label{fig2}
\end{figure*}

\section{Preliminary}

\subsection{Problem Formulation}

In a meta-learning paradigm of few-shot text classification, the source classes $Y_{train}$ are divided into sets of meta-tasks (or episodes), the model is trained with numerous episodes and it is evaluated on target classes $Y_{test}$, which are unseen during training. In general, meta-learning contains two phases: meta-training and meta-testing.

In a training episode, $N$ classes are sampled from training data $Y_{train}$. For each of these $N$ classes, $K$ labeled examples are sampled as the support set $S$ and another $M$ examples as the query set $Q$, donated as $S = {(X_i, Y_i)}^{N \times K} _{i=1}$ and $Q = {(X_j, Y_j)}^{N \times M} _{j=1}$. The model makes predictions about the query set $Q$ based on the given support set $S$,\textbf{ called the $N$-way $K$-shot task}. And then the model updates the parameters with the prediction results, which will be repeated multiple times.

% After finishing meta-training, the performance of the model will be evaluated by the same episode-based mechanism. 
In a testing episode, $N$ novel classes will be sampled from $Y_{test}$, which is disjoint to $Y_{train}$. Then the support set $S$ and the query set $Q$ will be sampled from the $N$ classes like in meta-training. The performance of the model will be evaluated through the average classification accuracy on the query set $Q$ across all testing episodes.

\subsection{Prompting and Feature Encoding}

The semantic information of the class label and the sample are naturally correlative. To make the correlation between samples and the label semantics, we use the encoding paradigm of prompt learning.
% We first prompt the support sample and query sample sentences. This is because that each sentence contains a limited number of words, through prompting the sentence with a template can get more class-related information. And prompting can construct the connection between the representation of the sentence and the representation of the label.
Our prompt templates are automatically selected from the OpenPrompt~\citep{Openprompt} knowledge base for different classification tasks. For example, to formulate the news classification task, we use a preset selector to select the candidate template and concatenate it with the sentence. The original input x with a selected template $T(\cdot)$ = This is a \texttt{[MASK]} news: to get the prompt input $T(x)$ = This is a \texttt{[MASK]} news: x. The prompted sentences are then put into a BERT encoder. Through BERT~\citep{Devlin_2019}, we get the token at the \texttt{[MASK]} position of the last layer as a sentence representation, 
\begin{equation}
\mathbf{v} = f_{\Phi}(T(x)) = \mathbf{h_{\texttt{[MASK]}}}.
\label{eq.sent}
\end{equation}

We also enter each label name (from datasets) into the BERT encoder to get the token of the label name, which is used as the label representation. If the label name is broken down into multiple words, the average of them is used,
\begin{equation}
\mathbf{u} = f_{\Phi}(label) = Mean(\mathbf{h}).
\label{eq.label}
\end{equation}

\subsection{Prototypical Networks}

Considering the wide application and powerful performance of the Prototype Networks (PN) \citep{proto_2017}, we combine our strategy with it to conduct experiments to verify performance. And our strategy can be also incorporated into other meta-learners, like RRML~\citep{BertinettoHTV19}.

% We choose Prototypical Networks as the final classifier, because it is a non-parametric classifier, our method can be more focused on enabling feature encoders to generate sample representation similar to its label representation.

The core idea of Prototypical Networks is to learn to classify the query samples by metric their similarities to support samples in the representation space. Specifically, for each testing task, the prototype $\mathbf{c}_{k}$ of the $k$-th class is obtained by averaging $K$ mapped support samples of the $k$-th class,

\begin{equation}
    \mathbf{c}_{k} = \frac{1}{K} \sum^K_{i=0} \mathbf{v}^k ,
    \label{eq.pro}
\end{equation}

And for a query sample representation $\mathbf{v}$, we get its probability score for class by,
\begin{equation}
    P_{\mathcal{M}}\left(y_{k} \mid x\right)=\frac{\exp S\left(\mathbf{v}^q_i, \mathbf{c}_{k}\right)}{\sum_{k \prime} \exp S\left(\mathbf{v}^q_i, \mathbf{c}_{k^{\prime}}\right)},
    \label{eq5}
\end{equation}
where $S(\cdot)$ is the cosine similarity.

Then we make prediction by ${\arg \max}$ function,
\begin{equation}
    \widetilde{y}= \mathop{\arg \max}\limits_{k} P_{\mathcal{M}}\left(y_{k} \mid x \right).
    \label{eq6}
\end{equation}

\section{Label-guided Distance Scaling}

Our proposed label-guided distance scaling (LDS) strategy is constructed to learn discriminative text representations guided by label semantic information. Fig.~\ref{fig2} gives an overview of our method. Firstly, the input sentences are prompted using the prompt templates, and the sample representations and label representations are obtained by a BERT encoder. Secondly, in the training stage, the label-guided loss is used to pull in the distance between each sample representation and its corresponding label representation. Finally, in the testing stage, Label-guided Scaler mixes support sample representations and label representations, pulling support sample representations further closer to the class centers. Classification can be done by combining with basic meta-learners, like Prototype Networks. The overall training procedure of LDS is summarized in Algorithm~\ref{alg:algorithm} in the Appendix \ref{sec:appendix}. 

\subsection{Distance Scaling in the training stage}

To strengthen the correlation between samples and the label semantics in the training stage, we inject the label semantics into samples through the training objective. 
However, most previously designed losses ignore the label semantics. Specifically, cross-entropy loss can not cluster the same class samples well. Contrastive loss tries to maximize intra-class similarity and minimize inter-class similarity, with no class centers to guide classification. 
Hence, we design a label-guided loss that pulls in the distance between each sample representation and its corresponding label representation, while pushing out the distance between it and other label representations. 
By this training objective, the labels can be regarded as the class centers.

For each episode, we combine the $N \times K$ samples ${x^s_i}$ in support set $S$ and the $N \times M$ samples ${x^q_j}$ in query set $Q$ as a training batch $B = \{x_1,x_2,\dots, x_{N(K+M)}\}$. For each $x_t \in B$, we denote its label as $y_t$, denote its label representation as $\mathbf{u}_t$, and denote its sample representation as $\mathbf{v}_t$. We construct the instance-label pairs and minimize the following loss function,

\begin{footnotesize}
\begin{equation}
\mathcal{L}_{LG}\hspace{-1mm}=\hspace{-1mm}-\frac{1}{c}\hspace{-1mm}\sum\limits_{x \in B} \log\frac{\sum\limits_{y_r=y_t}\hspace{-2mm}\exp(\mathbf{v}_t\cdot \mathbf{u}_r/\tau)}{\sum\limits_{y_r=y_t}\hspace{-2mm}\exp(\mathbf{v}_t \hspace{-1mm} \cdot \mathbf{u}_r/\tau)\hspace{-1mm}+\hspace{-3mm}\sum\limits_{y_{r \prime}\neq y_t}\hspace{-2mm}\exp(\mathbf{v}_t \hspace{-1mm} \cdot \mathbf{u}_{r \prime} /\tau)},
\label{eq:LG}
\end{equation}
\end{footnotesize}
where $c$ is the number of samples in the batch and $\tau$ is a temperature factor that scales the inner products. The inner product is used because the inner product not only calculates the difference in angle between two vectors but also takes into account the similarity in length between two vectors, which allows the sample representation to more closely resemble its label representation.

We further construct the label-label pairs to make label representations distinguishable, and minimize the following loss function for regularization during training,

\begin{footnotesize}
\begin{equation}
    \mathcal{L}_{label}=\hspace{-1mm}- \frac{1}{N} \hspace{-1mm}\sum\limits_{i=1}^{N}\log\frac{\exp(\mathbf{u}_i\cdot \mathbf{u}_i/\tau)}{\exp(\mathbf{u}_i\cdot \mathbf{u}_i/\tau)+\sum\limits_{i \neq j }\exp(\mathbf{u}_i\cdot \mathbf{u}_j /\tau)}.
\label{eq3}
\end{equation}
\end{footnotesize}

The overall objective is to minimize the sum of the two losses,
\begin{equation}
    \mathcal{L}_{all}=\mathcal{L}_{LG} + \mathcal{L}_{label}.
    \label{eq4}
\end{equation}

\subsection{Label-guided Scaler in the testing stage}

Considering that the support samples are randomly selected, the representation of these samples may deviate from the center of the class distribution. This discrepancy can result in query samples having closer proximity to support samples from a different class than to those of the same class, leading to misclassifications. To address this issue, we estimate the class centers with the aid of the labels. We formulate the estimation as an Expectation Maximization (EM) algorithm. In this formulation, we treat each support sample as a random variable representing the original estimated mean of the distribution. Since the labels of the support set samples are known and labels adhere to the class distribution, we scale each original support sample representation to align with its label representation by EM. This approach ensures a more accurate alignment of support samples with their respective class distributions, effectively mitigating misclassifications induced by the initial randomness in support sample selection.

Specifically, we combine the $K$ support sample representations with the $K$ corresponding label representations as $C = \{\mathbf s_1,...,\mathbf s_K, \mathbf s_{11},...\mathbf s_{1K}\}$ and perform E-step and M-step.

\textbf{E-step (Expectation)}: We calculate posterior probabilities by utilizing the current Gaussian Mixture Model (GMM) parameters to compute the posterior probability of the support sample belonging to its class center,
\begin{equation}
    P(z_i|\mathbf{s}_i) = \frac{w_i \cdot \mathcal{N}(\mathbf{s}_i|\mu_i, \Sigma_i)}{\sum_{j=1} w_j \cdot \mathcal{N}(\mathbf{s}_j|\mu_j,\Sigma_j)},
    \label{eq.6}
\end{equation}
where $\mathbf{s}_i$ is $i$th sample representation in $C$, $w_i$ is the weight of the component, $\mu_i$ is corresponding the mean and $\Sigma_i$ is the covariance matrix.

\textbf{M-Step (Maximization)}: The parameters of the GMM $\theta = \{\mu, w, \Sigma \}$are updated based on the computed posterior probabilities, repeat iterations until convergence.

\begin{footnotesize}
\begin{subequations}
\begin{align}
\mu_k^{t+1} &= \frac{\sum^{|C|}_{i=1}P(z_i=k|\mathbf{s}_i,\theta^{t}) \cdot \mathbf{s}_i}{\sum^{|C|}_{i=1}P(z_i=k|\mathbf{s}_i, \theta^{t})}, \\
\Sigma_k^{t+1} &= \frac{\sum^{|C|}_{i=1}\hspace{-1mm}P(z_i=k|\mathbf{s}_i,\theta^{t})\hspace{-1mm} \cdot \hspace{-1mm}(\mathbf{s}_i \hspace{-1mm} -\mu_i^{t+1}) \hspace{-1mm} \cdot \hspace{-1mm} (\mathbf{s}_i \hspace{-1mm} -\mu_i^{t+1})^T }{\sum^{|C|}_{i=1}P(z_i=k|\mathbf{s}_i, \theta^{t})}, \\
w_k^{t+1} &= \frac{\sum_i P(z_i=k|\mathbf{s}_i,\theta^{t})}{|C|}.
\end{align}
\label{eq.7}
\end{subequations}  
\end{footnotesize}

Here, we use the updated $w_i$ as the weight of the $i$th support sample representation and $w_{1i}$ as the weight of its label representation to enhance each support sample representation $\mathbf{s}_i$ with its label representation $\mathbf{s}_{1i}$, following,

\begin{equation}
    \mathbf{v}^s_i = \frac{w_i \cdot \mathbf{s}_i + w_{1i} \cdot \mathbf{s}_{1i}}{w_{i}+w_{1i}}.
    \label{fuse}
\end{equation}

% \begin{table*}[]
% \centering

% \begin{tabular}{lcccc}
% \toprule[1pt]
% \textbf{Dataset} & \textbf{Avg. text length} & \textbf{Samples per class} & \textbf{Total samples} & \textbf{Train / Valid / Test (Total) classes} \\ \hline
% HuffPost~\citep{Bao_2019} & 11.48 & 900  & 36900 & 20 / 5 / 16 (41)\\
% Amazon~\citep{amazon} & 143.46 & 1000  & 24000 & 10 / 5 / 9 (24) \\
% Reuters~\citep{Bao_2019} & 181.41 & 20 & 620 & 15 / 5 / 11 (31) \\
% 20News~\citep{20news} & 279.32 & 941 & 18828 & 8 / 5 / 7 (20) \\ \hline
% Banking77~\citep{Casanueva_Temčinas_Gerz_Henderson_Vulić_2020} & 11.77 & 170  & 13083  & 30 / 15 / 32 (77) \\
% Clinc150~\citep{Larson_Mahendran_Peper_Clarke_Lee_Hill_Kummerfeld_Leach_Laurenzano_Tang_2019} & 8.31 & 150  & 22500  & 60 / 15 / 75 (150) \\ \bottomrule[1pt]
% \end{tabular}
% \caption{The statistics of few-shot text classification datasets.}
% \label{tb:dataset}
% \end{table*}

\section{Experiments}
\subsection{Datasets}

Following \citep{TART}, we evaluate our LDS-PN model under typical 5-way tasks on four news or review classification datasets: \textbf{HuffPost}~\citep{Bao_2019}, \textbf{Amazon}~\citep{amazon}, \textbf{Reuters}~\citep{Bao_2019}, and \textbf{20News}~\citep{20news}. Additionally, we further explore its performance under 10-way and 15-way tasks in two intent detection datasets: \textbf{Banking77}~\citep{Casanueva_Temčinas_Gerz_Henderson_Vulić_2020}, and \textbf{Clinic150}~\citep{Larson_Mahendran_Peper_Clarke_Lee_Hill_Kummerfeld_Leach_Laurenzano_Tang_2019}. The average length of sentences in news or review classification datasets is much longer than those in intent detection datasets. The statistics of the datasets are shown in Table~\ref{tb:dataset}.

\begin{table}[!h]
\setlength{\tabcolsep}{5pt}
\begin{tabular}{lccc}
\toprule[1pt]
\small \textbf{Dataset}   & \small \textbf{Avg. Len} & \small \textbf{Samples} & \small \textbf{Train / Valid / Test} \\ \hline
\small HuffPost  & 11.48    & 36900      & 20 / 5 / 16     \\
\small Amazon    & 143.46   & 24000      & 10 / 5 / 9      \\
\small Reuters   & 181.41   & 620        & 15 / 5 / 11     \\
\small 20News    & 279.32   & 18828      & 8 / 5 / 7       \\ \hline
\small Banking77 & 11.77    & 13083      & 30 / 15 / 32    \\
\small Clinc150  & 8.31     & 22500      & 60 / 15 / 75    \\ 
\bottomrule[1pt]
\end{tabular}
\caption{Dataset statistics.}
\label{tb:dataset}
\end{table}

\begin{table*}[h]
\setlength{\tabcolsep}{3.5pt}
\centering
\begin{tabular}{lcccccccccc}
\toprule
\multirow{2}{*}{\textbf{Methods}} & \multicolumn{2}{c}{\textbf{HuffPost}} & \multicolumn{2}{c}{\textbf{Amazon}} & \multicolumn{2}{c}{\textbf{Reuters}} & \multicolumn{2}{c}{\textbf{20News}} & \multicolumn{2}{c}{\textbf{Average}} \\ \cline{2-11} 
 & 1-shot & 5-shot & 1-shot & 5-shot & 1-shot & 5-shot & 1-shot & 5-shot & 1-shot & 5-shot \\ \hline
PN (NeurIPS 17) & 35.7 & 41.3 & 37.6 & 52.1 & 59.6 & 66.9 & 37.8 & 45.3 & 42.7 & 51.4 \\
MAML (ICML 17) & 35.9 & 49.3 & 39.6 & 47.1 & 54.6 & 62.9 & 33.8 & 43.7 & 40.9 & 50.8 \\
IN (EMNLP 19) & 38.7 & 49.1 & 34.9 & 41.3 & 59.4 & 67.9 & 28.7 & 33.3 & 40.4 & 47.9 \\
DS-FSL (ICLR 20) & 43.0 & 63.5 & 62.6 & 81.1 & 81.8 & 96.0 & 52.1 & 68.3 & 59.9 & 77.2 \\
LaSAML (ACL 21) & 62.2 & 70.1 & 62.2 & 79.1 & 90.0 & 96.7 & 56.2 & 77.7 & 67.7 & 80.9 \\
MLADA (ACL 21) & 45.0 & 64.9 & 68.4 & 86.0 & 82.3 & 96.7 & 59.6 & 77.8 & 63.8 & 81.4 \\
ContrastNet (AAAI 22) & 51.8 & 67.8 & 73.5 & 83.6 & 88.5 & 94.6 & 70.9 & 80.5 & 71.2 & 81.6 \\
ProtoVerb (ACL 22) & 53.1 & 70.8 & 72.4 & 84.7 & 85.4 & 94.2 & 60.2 & 83.1 & 67.8 & 83.2 \\
Shot-DE (AAAI 23) & 51.9 & 71.4 & 76.1 & 86.9 & 90.6 & 95.1 & 71.0 & 83.2 & 72.4 & 84.2 \\
Way-DE (AAAI 23) & 51.9 & 71.7 & 76.1 & \textbf{87.4} & 90.6 & 95.2 & 71.0 & 83.2 & 72.4 & 84.4 \\
% Meta-SN (2023) & 63.1 & 71.3 & 77.5 & \textbf{89.1} & 87.9 & 96.7 & 72.1 & 83.2 & 75.2 & 85.1 \\
TART (ACL 23) & 46.5 & 68.9 & 73.7 & 84.3 & 86.9 & 95.6 & 73.2 & 84.9 & 70.1 & 83.4 \\ 
SPCNet (TIS 24) &53.3 & 65.5 & 76.3 &85.3 &86.8 &95.4  &72.3 &81.7 &72.2 &82.0 \\
\hline
LDS-PN (Ours) & \textbf{67.5} & \textbf{76.4} & \textbf{81.8} & 85.1 & \textbf{92.4} & \textbf{97.9} & \textbf{75.3} & \textbf{85.4} & \textbf{81.8} & \textbf{86.8} \\ \bottomrule
\end{tabular}
\caption{The 5-way 1-shot and 5-shot average accuracy on news or review classification datasets.}
\label{tb:news}
\end{table*}

\begin{table*}[]
\setlength{\tabcolsep}{3.5pt}
\centering
\begin{tabular}{lcccccccccc}
\toprule
\multirow{3}{*}{\textbf{Methods}} & \multicolumn{4}{c}{\textbf{Banking77}} & \multicolumn{4}{c}{\textbf{Clinc150 (cross-domain)}} & \multicolumn{2}{c}{\textbf{Average}} \\ \cline{2-11} 
 & \multicolumn{2}{c}{10-way} & \multicolumn{2}{c}{15-way} & \multicolumn{2}{c}{10-way} & \multicolumn{2}{c}{15-way} & - & - \\ \cline{2-11} 
 & 1-shot & 5-shot & 1-shot & 5-shot & 1-shot & 5-shot & 1-shot & 5-shot & 1-shot & 5-shot \\ \hline
PN (NeurIPS 17) & 40.9 & 54.6 & 32.5 & 46.1 & 42.6 & 57.7 & 34.2 & 48.3 & 37.6 & 51.7 \\
DS-FSL (ICLR 20) & 59.3 & 83.7 & 53.4 & 79.0 & 55.6 & 78.8 & 53.4 & 79.7 & 55.4 & 80.3 \\
LaSAML (ACL 21) & 82.8 & 88.1 & 78.8 & 84.4 & 77.6 & 88.3 & 72.5 & 84.9 & 77.9 & 86.4 \\
MLADA (ACL 21) & 61.2 & 80.8 & 55.6 & 74.8 & 65.3 & 85.5 & 51.8 & 77.8 & 58.5 & 79.7 \\
ContrastNet (AAAI 22) & 75.5 & 87.9 & 69.9 & 83.6 & 81.9 & 93.1 & 75.2 & 90.7 & 75.6 & 88.8 \\
ProtoVerb (ACL 22) & 72.3 & 87.0 & 66.6 & 83.2 & 87.6 & 92.4 & 84.4 & 90.0 & 77.7 & 88.2 \\ 
AMGS (COlING 22) & 71.4 & 88.8 & 63.6 & 84.9 & 69.2 & 88.3 & 62.1 & 84.1 & 66.6 & 86.5 \\ 
% Shot-DE (AAAI 2023) &  77.3 & 90.3 & 74.9 & 86.5 & 89.6 & 95.1 & 86.1 & 93.6 & 81.9 & 91.4 \\
% Way-DE (AAAI 2023) & 77.3 & 90.3 & 74.9 & 86.5 & 89.6 & 95.1 & 86.1 & 93.6 & 81.9 & 91.4 \\
TART (ACL 23) & 56.1 & 66.2 & 55.6 & 62.3 & 72.4 & 88.9 & 67.5 & 85.2 & 62.9 & 75.7 \\ 
DCCL (CSCWD 24) & - &  88.5 & -& 85.8  & - & 88.3 & - &  84.2 & - & 86.7 \\
\hline
LDS-PN (Ours) & \textbf{85.8} &\textbf{89.4} & \textbf{81.0} & \textbf{85.9} & \textbf{93.1} & \textbf{93.5} & \textbf{91.3} & \textbf{92.2} & \textbf{87.8} & \textbf{90.3} \\ \bottomrule
\end{tabular}
\caption{The 10-way 1-shot, 10-way 5-shot, 15-way 1-shot, and 15-way 5-shot average accuracy on intent classification datasets. "-" is because that DCCL is only designed for 5\&10-shot tasks.}
\label{tb:intent}
\end{table*}

\subsection{Baselines}

We compare our approach against well-established few-shot baselines: 
\textbf{PN}~\citep{proto_2017},
\textbf{MAML}~\citep{MAML_2017}, 
\textbf{IN}~\citep{Geng2019}, 
\textbf{DS-FSL}~\citep{Bao_2019},
\textbf{LaSAML}~\citep{Luo_2021},
\textbf{MLADA}~\citep{Han__2021},
\textbf{AMGS}~\citep{Lei_Hu_Luo_Peng_Wang_2022},
\textbf{ContrastNet}~\citep{Chen_Zhang_Mao_Xue_2022},
\textbf{ProtoVerb}~\citep{proto_22},
\textbf{DE}~\citep{Liu2023},
\textbf{TART}~\citep{TART},
\textbf{DCCL}~\citep{CSCWD},
\textbf{SPCNet}~\citep{SPContrastNet}.
Their detail descriptions are in Appendix \ref{sec.baselines}.

\subsection{Implementation Details}
\textbf{Evaluation Metric} We follow~\citep{Bao_2019}, utilizing the identical dataset division and evaluating performance based on accuracy (ACC). In our experiments, we randomly sample 100, 100, and 1000 task episodes for each training, validation, and testing epoch, respectively. All reported results are the average ACC across 5 runs.

\noindent \textbf{Parameter Setting} We adopt the Adam~\citep{Kingma_Ba_2014} algorithm with a learning rate of 1e-6 as the optimizer. We take the results of baseline methods from~\citep{Liu2023, TART} for typical 5-way classification, and from~\citep{Luo_2021,Lei_Hu_Luo_Peng_Wang_2022} for 10\&15-way classification. Specific parameter settings can be found in our publicly available repository \footnote{https://anonymous.4open.science/r/Label-guided-Text-Classification}. All the experiments are conducted with NVIDIA RTX A4000 GPUs (15 epochs per hour).

\subsection{Typical 5-way Task Results}

Typical few-shot text classification algorithms are tested on 5-way 1-shot and 5-way 5-shot tasks. We conduct experiments on four common news or review datasets. The main experiment results are displayed in Table~\ref{tb:news}. From the result, we can make the following observation: (1) The proposed LDS-PN achieves significant performance improvement over the vanilla PN 39.1\%, 35.4\% in 1-shot and 5-shot scenarios. (2) LDS-PN achieves significant performance improvement over other baselines across all the datasets. Specifically, in the 1-shot and 5-shot scenarios, LDS-PN is averagely 9.4\% and 2.4\% better than the current state-of-the-art method Way-DE~\citep{Liu2023}. This is because LDS-PN makes the class distribution more distinguishable and designates explicit class centers, which helps to generate a more reliable prototype. (3) The 5-shot result on Amazon dataset is suboptimal. This is because the Amazon dataset contains noisy reviews (1000 per class), and the label names may differ from the actual class centers. Although our LDS-PN does not exceed several strong baselines on the Amazon dataset in the 5-shot scenario, it is still best in the 1-shot scenario.

\subsection{Further 10\&15-way Task Results}
We further compare LDS-PN with existing methods under much harder 10-way and 15-way tasks on Banking77 and Clinic150 datasets. Table~\ref{tb:intent} displays the results. It can be seen that LDS-PN still achieves the best compared with the baseline methods.On average, LDS-PN is 50.2\% and 38.6\% better than vanilla PN. Moreover, LDS-PN has an average improvement of 10.1\% and 2.1\% over the second best in the 1-shot and 5-shot scenarios.

\subsection{Ablation Study}

In this section, we build the experiment to explore 4 parts. The first is the impact of Distance Scaling in the training stage. The second is the effectiveness of the Label-guided Scaler in the testing stage. The third is the effect of combining our method with other meta-learning classifiers, proving the universality of our method. Finally, we explore the impact of prompt learning on our strategy, the results are in the Appendix \ref{sec.pt}.

\begin{table}[h]
% \resizebox{\linewidth}{!}{
\centering

\setlength{\tabcolsep}{4.5pt}
\begin{tabular}{|l|l|cc|cc|}
\hline
\multicolumn{1}{|c|}{\multirow{3}{*}{\textbf{\small Methods}}} & \multicolumn{1}{c|}{\multirow{3}{*}{\textbf{ \small Loss}}} & \multicolumn{2}{c|}{\textbf{ \small HuffPost}} & \multicolumn{2}{c|}{\textbf{\small Clinc150}} \\ \cline{3-6} 
\multicolumn{1}{|c|}{} & \multicolumn{1}{c|}{} & \multicolumn{2}{c|}{5-way} & \multicolumn{2}{c|}{10-way} \\ \cline{3-6} 
\multicolumn{1}{|c|}{} & \multicolumn{1}{c|}{} & \multicolumn{1}{c|}{1-shot} & 5-shot & \multicolumn{1}{c|}{1-shot} & 5-shot \\ \hline
\multirow{2}{*}{\small \begin{tabular}[c]{@{}l@{}}LDS-PN\\ w/o DS\end{tabular}} & $\mathcal{L}_{CE}$ & \multicolumn{1}{c|}{55.6} & 69.5 & \multicolumn{1}{c|}{83.3} & 86.4 \\ \cline{2-6} 
 &$\mathcal{L}_{SC}$ & \multicolumn{1}{c|}{53.1} & 70.8 & \multicolumn{1}{c|}{87.6} & 92.4 \\ \hline
\multirow{2}{*}{\small \begin{tabular}[c]{@{}l@{}}LDS-PN\\ w DS\end{tabular}}  & $\mathcal{L}_{LG}$ & \multicolumn{1}{c|}{65.3} & 76.1 & \multicolumn{1}{c|}{93.1} & 93.4 \\ \cline{2-6} 
 & $\mathcal{L}_{all}$ & \multicolumn{1}{c|}{\textbf{67.5}} & \textbf{76.4} & \multicolumn{1}{c|}{\textbf{93.1}} & \textbf{93.5} \\ \hline
\end{tabular}
% }
\caption{Ablation study results of the Distance Scaling in the training stage (DS). $\mathcal{L}_{all} = \mathcal{L}_{LG} + \mathcal{L}_{label}$.}
\label{tb:loss}
\end{table}

\subsubsection{LDS with vs without Distance Scaling in the training stage}

We explore the effect of our designed \textbf{D}istance \textbf{S}caling in the training stage (DS) by comparing LDS-PN with DS and without DS. Without DS, we choose two loss functions: Cross-entropy loss $\mathcal{L}_{CE}$ is commonly used in prompt learning, and supervised contrastive loss $\mathcal{L}_{SC}$ is used in ProtoVerb. With DS, the loss functions are $\mathcal{L}_{LG} + \mathcal{L}_{label}$ with regularization, and  $\mathcal{L}_{LG}$ without regularization. Table~\ref{tb:loss} shows the results on the HuffPost and clinc150 datasets. It can be seen that LDS-PN with $\mathcal{L}_{all}$ has the best scores, showing the effectiveness of the label-guided distance scaling training stage. It is easy to understand that without DS, the encoder can not generate distinguishable sample representations. Specifically, (1) $\mathcal{L}_{CE}$ loss maximizes the probability of correct classification, so it can only constrain the representation of samples of the same class to be close to each other, but it cannot constrain the representation of different samples to be far together. (2) $\mathcal{L}_{SC}$ loss can pull in the same class samples and push away different class samples, but it ignores the guidance of label semantic information, so it cannot solve the prediction conflict that the support samples are randomly selected and not representative. DS($\mathcal{L}_{all}$) can pull in the distance between each sample representation and the corresponding class label representation while pushing away the different class label representations, which makes the sample representation easier to distinguish, and under this optimization goal support sample representations can further close to the class center in the testing stage by using the label representations.

\begin{table}[]
\centering
% \resizebox{\linewidth}{!}{

\setlength{\tabcolsep}{3pt}
\begin{tabular}{|l|l|cc|cc|}
\hline
\multicolumn{1}{|c|}{\multirow{3}{*}{\textbf{\small Methods}}} & \multirow{3}{*}{\textbf{\small Operation}} & \multicolumn{2}{c|}{\textbf{\small HuffPost}} & \multicolumn{2}{c|}{\textbf{ \small Clinc150}} \\ \cline{3-6} 
\multicolumn{1}{|c|}{} &  & \multicolumn{2}{c|}{5-way} & \multicolumn{2}{c|}{10-way} \\ \cline{3-6} 
\multicolumn{1}{|c|}{} &  & \multicolumn{1}{c|}{1-shot} & 5-shot & \multicolumn{1}{c|}{1-shot} & 5-shot \\ \hline
\small  w/o LS & \small None & \multicolumn{1}{c|}{57.3} & 72.9 & \multicolumn{1}{c|}{84.8} & 93.2 \\ \hline
\multirow{3}{*}{\small w LS}  & \small Connect & \multicolumn{1}{c|}{43.2} & 48.6 & \multicolumn{1}{c|}{63.6} & 54.5 \\  \cline{2-6} 
 & \small Attention & \multicolumn{1}{c|}{52.7} & 73.0 & \multicolumn{1}{c|}{89.3} & 93.0 \\ \cline{2-6} 
 &  \small EM & \multicolumn{1}{c|}{\textbf{67.5}} & \textbf{76.4} & \multicolumn{1}{c|}{\textbf{93.1}} & \textbf{93.5 }\\
 \hline
\end{tabular}
% }
\caption{Ablation study results of the Label-guided Scaler in the testing stage (LS).}
\label{tb:LG}
\end{table}

\subsubsection{LDS with vs without Label-guided Scaler in the testing stage}

We investigate the effectiveness of the \textbf{L}abel-guided \textbf{S}caler in the testing stage (LS) on HuffPost and Clinic150 datasets. As shown in Table~\ref{tb:LG}, (1) compared with LS using the EM algorithm in this paper, LDS-PN without LS drops average 9.2\% in the 1-shot task, which verifies the effectiveness of the Label-guided Scaler in the testing stage. This is because LS can utilize the label semantic information to further enhance the support sample representations.
(2) Compared to several different scaling methods for LS, it can be found that LS using the EM algorithm has the best results. The reason is that attention and connect layers need training and they easily trap in overfitting because of the limited samples. Differently, the EM algorithm is non-parametric, and it mixes the information of both labels and samples, pulling sample representations closer to the class centers effectively.

\begin{figure}[h]
\centering
\subfigure[LDS w/o LS]{\includegraphics[width=.34\textwidth]{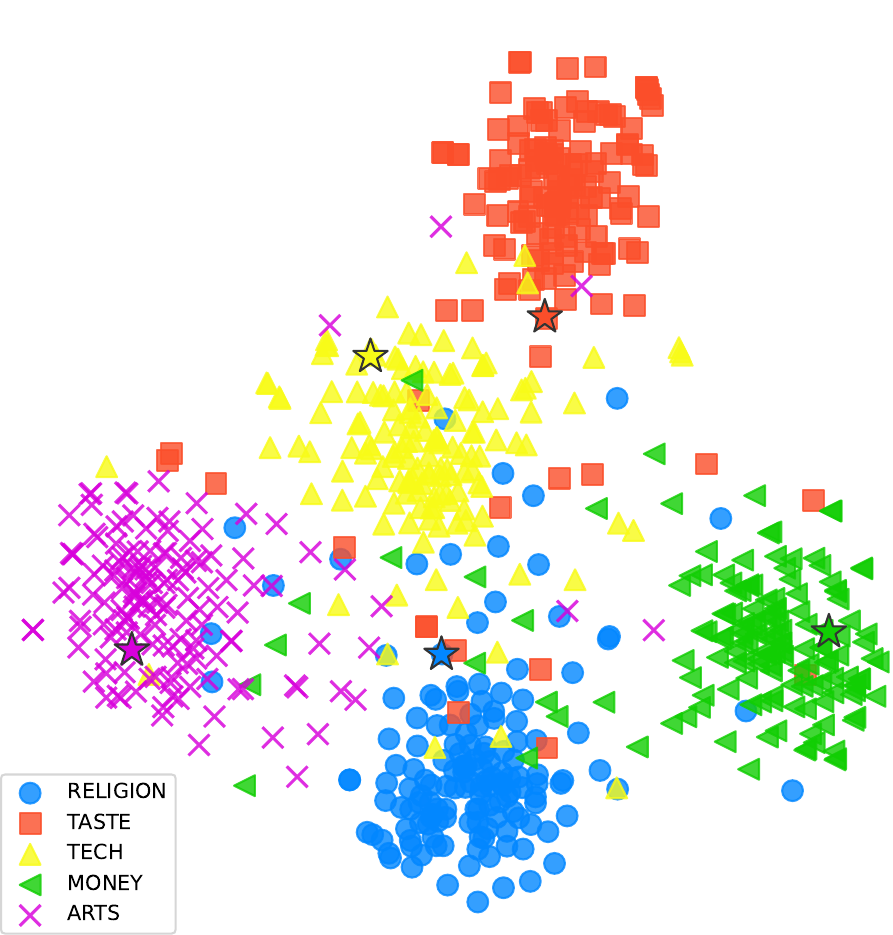}
\label{casea}
}

\subfigure[LDS (ours)]{\includegraphics[width=.34\textwidth]{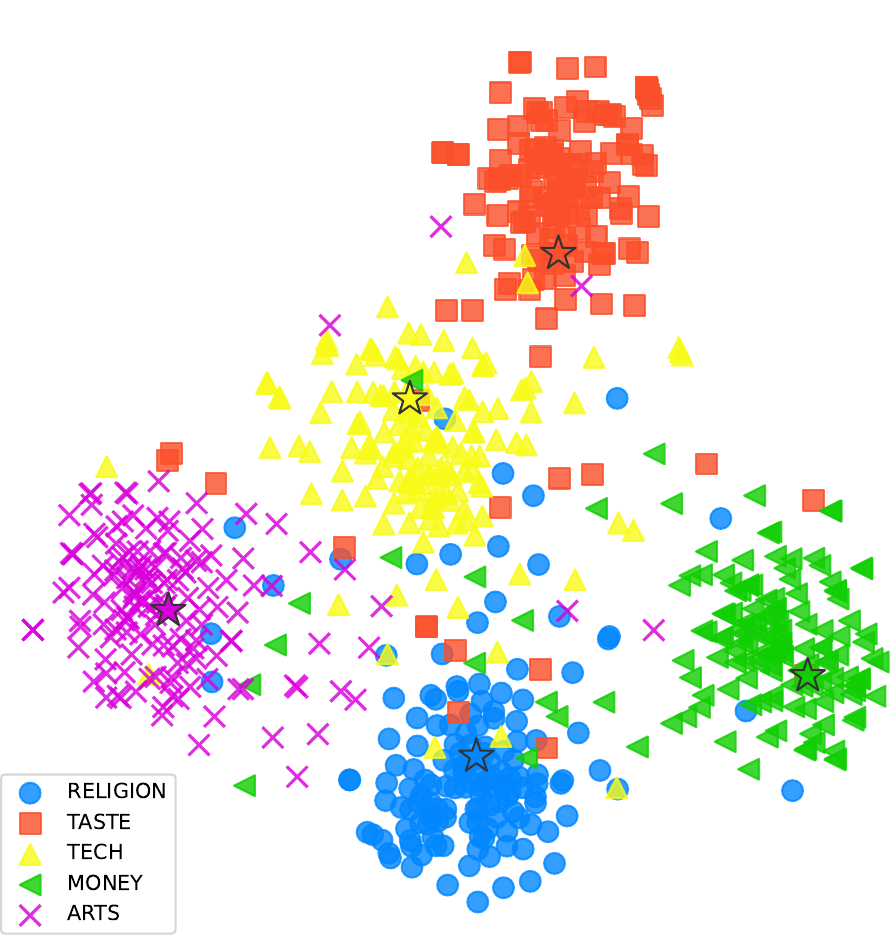}
\label{caseb}
}
\caption{Visualization of text representations from HuffPost dataset, given by (a) LDS w/o LS and (b) LDS (ours). The pentagrams represent the support samples.}
\label{fig.case}
\end{figure}

\begin{table}[h]
\centering

\begin{tabular}{|l|ll|ll|}
\hline
\multicolumn{1}{|c|}{\multirow{3}{*}{\textbf{Methods}}} & \multicolumn{2}{c|}{\textbf{ HuffPost}}        & \multicolumn{2}{c|}{\textbf{ Clinc150}}        \\ \cline{2-5} 
                         & \multicolumn{2}{c|}{5-way}           & \multicolumn{2}{c|}{10-way}          \\ \cline{2-5} 
                         & \multicolumn{1}{c|}{1-shot} & 5-shot & \multicolumn{1}{c|}{1-shot} & 5-shot \\ \hline
\small BERT-PN                  & \multicolumn{1}{c|}{46.1}   & 65.6   & \multicolumn{1}{c|}{63.0}   & 81.6   \\ 
\small LDS-PN                   & \multicolumn{1}{c|}{\textbf{67.5}} &  \textbf{76.4} & \multicolumn{1}{c|}{\textbf{93.1}} & \textbf{93.5}         \\ \hline
\small BERT-RRML                & \multicolumn{1}{c|}{40.8}   & 62.0   & \multicolumn{1}{c|}{57.6}   & 80.8   \\ 
\small LDS-RRML                 & \multicolumn{1}{c|}{\textbf{67.0}}       &    \textbf{70.8}    & \multicolumn{1}{c|}{\textbf{93.3} }      &      \textbf{93.8}  \\ \hline
\end{tabular}

\caption{Ablation study results of the integrating LDS with RRML.}
\label{tb:cl}
\end{table}

% \begin{figure}[h]
% \centering
% \subfigure[ProtoVerb]{\includegraphics[width=.34\textwidth]{figure/baseline_show.pdf}
% \label{visb}}
% \subfigure[LDS (ours)]{\includegraphics[width=.34\textwidth]{figure/ours_show.pdf}
% \label{visc}}
% \caption{Visualization of text representations from HuffPost dataset, given by (a) ProtoVerb and (b) LDS (ours). The pentagrams represent the class prototypes.}
% \label{figvis}
% \end{figure}

\subsubsection{LDS Boosting Other Meta-learner}
Our proposed LDS strategy could also be used to boost other meta-learners. To further explore the potential of LDS, we incorporate it into the Ridge Regression Meta-learner (RRML)~\citep{BertinettoHTV19} to perform experiments on the HuffPost and Clinic150 datasets. As shown in Table~\ref{tb:cl}, LDS-RRML achieves significant performance improvement over the BERT-RRML average of 21\%. On Clinc150 dataset, LDS-RRML even outperforms LDS-PN, demonstrating the universality of our LDS strategy and opening up the possibility of combining LDS with more meta-learners.

\subsection{Visualization and Case Study}
To demonstrate our main module Label-guided Scaler (LS) can mitigate the misclassification caused by randomly selected support samples,  we visualize the sample representations produced by LDS w/o LS, and LDS using t-SNE~\citep{Maaten2008VisualizingDU}. We randomly select 5 novel classes (150 samples per class) on HuffPost dataset, where 1 sample per class is selected as a support sample. As shown in Fig~\ref{casea} (LDS w/o LS), the support samples are far away from the class centers, but in Fig~\ref{caseb} (LDS), the support samples are pulled closer to the class centers. The specific cosine distances between support samples and their class centers are shown in Table~\ref{tb:case}. The above results demonstrate that LS in the testing stage can reduce unsatisfactory support sample representations, thereby reducing misclassification caused by randomly selected support samples.

\section{Conclusion}
We analyze the misclassification in the few-shot scenario caused by the randomly selected labeled samples.
Then, we propose a \textbf{L}abel-guided \textbf{D}istance \textbf{S}caling (LDS) strategy during both the training and the testing stages to reduce the prediction contradictions. 
Specifically, in the training stage, we design a label-guided loss to inject label semantics, making the sample representations closer to the corresponding label representations.
In the testing stage, we propose a Label-guided Scaler that scales support sample representations and label semantics, pulling support sample representations further closer to the class center. 
In this paper, we make a attempt to boost meta-learners by designing a strategy for the testing stage.

\section{Limitations}
Our approach is suitable for single-label classification tasks, but it may have a limitation in multi-label classification tasks. And our method introduces the label semantics, which takes up more resources during training, see Table \ref{tab:time} for details. When applying this research to real applications, our encoder and prompt may need to change to a more powerful model for better performance, see Appendix \ref{sec:LN} for details.

\bibliography{custom}

\appendix

\section{Method}
\subsection{Label Names Details}
\label{sec:LN}
In our paper, we don’t do any extra operation on labels, we merely use original label names from these datasets, and input them into the tokenizer, and then into the bert encoder, we use the mean of the $last\_hidden\_states$ as the label representations. We find that some label names are difficult to understand (e.g., in 20News, comp.os.ms-windows.misc), which is a limitation of our approach performance. However, our key idea is to use label semantics to fix the misclassification caused by randomly selected support samples, so we still use the original label names from the datasets, this can be future work to enhance the performance of LDS further. Since LDS relies on the label name selection, if enhancing the label names chosen like extending label names (e.g., KPT~\citep{KPT}, KPT++~\citep{KPT++}), or using a knowledge graph, the results of LDS will further improve.

\subsection{Pseudocode}
\label{sec:appendix}
Our approach follows the training strategy of meta-learning. Firstly, the source classes $Y_{train}$ are divided into sets of meta-tasks, including $N \times K$ sentences as support samples and $N \times M$ sentences as query samples. Secondly, the input sentences are prompted using the prompt templates, and the sample representations and label representations are obtained by a BERT encoder. Thirdly, Distance Scaling is used as an objective, to pull in the distance between each sample representation and its corresponding label representation. Unlike standard meta-learning, our losses do not depend on classification results. After training, we added a Label-guided Scaler in the testing stage to further guide the support samples, and to pull sample representations closer to their class centers.

\begin{algorithm}[tb]
\caption{Training procedure of LDS}
\label{alg:algorithm}
\textbf{Input}: Training data ${X_{train}, Y_{train}}$; T episodes and $ep$ epochs; $N$ classes in support set or query set; $K$ samples in each class in the support set and $M$ samples in each class in the query set; The parameters of feature encoder $\Phi$ \\
\textbf{Output}: Parameters $\Phi$ after training \\
\begin{algorithmic}[1] %[1] enables line numbers
\FOR{$i \in [1,ep]$} 

\STATE      $\mathcal{Y} \leftarrow  \Lambda(Y_{train}, N)$;   \textit{ //  select $N$ elements from $Y_{train}$ randomly.}
\FOR{$each j \in [1,T]$}
\STATE      $S, Q \leftarrow \emptyset, \emptyset$;
\FOR{$y \in \mathcal{Y}$}
\STATE        $ S \leftarrow S \cup \Lambda(X_{train}\{y\},K)$;
\STATE        $ Q \leftarrow Q \cup \Lambda(X_{train}\{y\} /\ S,M)$;
\STATE        $ LS \leftarrow  \Omega (S)$;     \textit{ //  get the class names}
\STATE        $ LQ \leftarrow  \Omega (Q)$;
\ENDFOR
\STATE  Input $S,Q,LS,LQ$ to the encoder\\
        get $\mathbf{v}^s, \mathbf{v}^q, \mathbf{u}^s, \mathbf{u}^q$ by Eq. \ref{eq.sent}, \ref{eq.label};
\STATE  Input $\mathbf{v}^s, \mathbf{u}^s$ to the label-guided scaler\\
        get final $\mathbf{v}^s$ by Eq. \ref{eq.6}, \ref{eq.7}, \ref{fuse};
\STATE  Update $\Phi$ by minimizing the loss of the Eq. \ref{eq4};
\ENDFOR
\ENDFOR
\end{algorithmic}
\end{algorithm}

\section{Experiments}

\subsection{Datasets}

Following \citep{TART}, we evaluate our LDS-PN model under typical 5-way tasks on four news or review classification datasets: \textbf{HuffPost}~\citep{Bao_2019}, \textbf{Amazon}~\citep{amazon}, \textbf{Reuters}~\citep{Bao_2019}, and \textbf{20News}~\citep{20news}. Additionally, we further explore its performance under 10-way and 15-way tasks in two intent detection datasets: \textbf{Banking77}~\citep{Casanueva_Temčinas_Gerz_Henderson_Vulić_2020}, and \textbf{Clinic150}~\citep{Larson_Mahendran_Peper_Clarke_Lee_Hill_Kummerfeld_Leach_Laurenzano_Tang_2019}. The average length of sentences in news or review classification datasets is much longer than those in intent detection datasets. The statistics of the datasets are shown in Table~\ref{tb:dataset}.

\textbf{HuffPost} comprises news headlines published on HuffPost spanning from 2012 to 2018. These headlines are categorized into 41 classes. Moreover, the sentences within this dataset tend to be shorter and less grammatically correct compared to formal phrases.

\textbf{Amazon} comprises product reviews spanning 1996 to 2014 across 24 product categories. Our objective is to classify the product categories of these reviews. Due to the considerable size of the original dataset, we employ a sampling strategy similar to that described in~\citep{Han__2021}, selecting a subset of 1000 reviews from each category for our analysis.

\textbf{Reuters} comprises concise articles from Reuters published in 1987. We utilize the standard ApteMod version of the dataset. Consistent with the methodology outlined in~\citep{Bao_2019}, we exclude multi-label articles and restrict our analysis to 31 classes, each containing a minimum of 20 articles.

\textbf{20News} encompasses 18,828 documents sourced from news discussion forums, categorized into 20 distinct topics. Notably, the lengths of sentences within this dataset exhibit considerable variation.

\textbf{Banking77} serves as a fine-grained single-domain dataset designed for intent detection. Within this dataset, certain categories exhibit similarity and may overlap with others. It encompasses 13,083 fine-grained intents distributed across 77 distinct classes within the banking domain.

\textbf{Clinc150} serves as a cross-domain intent classification dataset, featuring 150 intents spanning 10 distinct domains. It comprises 22,500 user utterances, with an even distribution across each intent, along with 1200 out-of-scope queries. For the purposes of this study, we disregard these out-of-scope examples. Consistent with~\citep{Luo_2021}, we partition the dataset into training, validation, and testing sets, allocating 4, 1, and 5 domains, respectively.

\subsection{Baselines}
\label{sec.baselines}
We compare our approach against several well-established few-shot baseline models: (1) \textbf{PN}~\citep{proto_2017} leverages Euclidean distance to measure query examples against the class vector averaged by support examples. (2) \textbf{MAML}~\citep{MAML_2017} trains a favorable initial point for the base learner by utilizing the meta-learning that learns among tasks. (3) \textbf{IN}~\citep{Geng2019} learns a generalized class-wise representation by leveraging a dynamic routing algorithm. (4) \textbf{DS-FSL}~\citep{Bao_2019} builds an attention generator to get the representations. (5) \textbf{LaSAML}~\citep{Luo_2021} extracts sentence representations from BERT with label names. (6) \textbf{MLADA}~\citep{Han__2021} introduces an adversarial domain adaptation network in meta-learning systems, which aims to generate generalized sample representations for new classes. (7) \textbf{AMGS}~\citep{Lei_Hu_Luo_Peng_Wang_2022} adds the Masked Language Modeling task as an auxiliary task and optimizes meta-learner via gradient similarity between it and the basic task. (8) \textbf{ContrastNet}~\citep{Chen_Zhang_Mao_Xue_2022} introduces instance-level and task-level regularization loss to learn sample representations. (9) \textbf{ProtoVerb}~\citep{proto_22} introduces instance-level and prototype-level contrastive loss in prompt learning to learn class prototypes from training instances. (10) \textbf{DE}~\citep{Liu2023} provides two strategies: Way-DE and Shot-DE to calibrate the data distribution by utilizing the top nearest query samples and generate more samples for the support set. (11) \textbf{TART}~\citep{TART} transforms the class prototypes to per-class fixed reference points in task-adaptive metric spaces.
(12) \textbf{DCCL}~\citep{CSCWD} introduces a distance coefficient to dynamically guide contrastive learning.
(13) \textbf{SPCNet}~\citep{SPContrastNet} combines an efficient self-paced episode sampling approach and contrastive learning.

\begin{table}[!h]
\centering
\begin{tabular}{|l|cc|cc|}
\hline
\multicolumn{1}{|c|}{\multirow{3}{*}{\textbf{Methods}}} & \multicolumn{2}{c|}{\textbf{HuffPost}}             & \multicolumn{2}{c|}{\textbf{Clinc150}}             \\ \cline{2-5} 
\multicolumn{1}{|c|}{} & \multicolumn{2}{c|}{5-way}       & \multicolumn{2}{c|}{10-way}      \\ \cline{2-5} 
\multicolumn{1}{|c|}{}                         & \multicolumn{1}{c|}{1-shot}        & 5-shot        & \multicolumn{1}{c|}{1-shot}        & 5-shot        \\ \hline
KNN                    & \multicolumn{1}{c|}{39.8} & 45.3 & \multicolumn{1}{c|}{78.1} & 74.0 \\ \hline
PN                     & \multicolumn{1}{c|}{39.8} & 49.2 & \multicolumn{1}{c|}{78.1} & 85.1 \\ \hline
PN+DS                  & \multicolumn{1}{c|}{57.3} & 72.9 & \multicolumn{1}{c|}{84.8} & 93.2 \\ \hline
PN+LS                  & \multicolumn{1}{c|}{55.6} & 69.5 & \multicolumn{1}{c|}{83.3} & 86.4 \\ \hline
LDS-PN                                         & \multicolumn{1}{c|}{\textbf{67.5}} & \textbf{76.4} & \multicolumn{1}{c|}{\textbf{93.1}} & \textbf{93.5} \\ \hline
\end{tabular}
\caption{Ablation study results of step-up from PN (baseline) to LDS-PN.}
\label{tb:step}
\end{table}

\begin{table*}[h]
\centering
\begin{tabular}{|l|l|r|r|r|}
\hline
\textbf{Methods} &
  \textbf{\begin{tabular}[c]{@{}l@{}}training time \\ per epoch(s)\end{tabular}} &
  \multicolumn{1}{l|}{\textbf{\begin{tabular}[c]{@{}l@{}}GPU memory \\ per training(M)\end{tabular}}} &
  \multicolumn{1}{l|}{\textbf{\begin{tabular}[c]{@{}l@{}}1000 times \\ inference time(s)\end{tabular}}} &
  \multicolumn{1}{l|}{\textbf{testing ACC}} \\ \hline
PN &
  53$(\pm 2.5)$ &
  7858 &
  250 &
  39.8 \\ \hline
LDS-PN &
  60$(\pm 2.2)$ &
  8167 &
  270 &
  67.5 \\ \hline
\end{tabular}
\caption{The results of comparison of computational resources and inference time on HuffPost dataset.}
\label{tab:time}
\end{table*}

\begin{table*}[h!]
\centering
\begin{tabular}{|l|l|cc|}
\hline
\multicolumn{1}{|c|}{\multirow{3}{*}{\textbf{Methods}}} & \multicolumn{1}{c|}{\multirow{3}{*}{\textbf{Prompt template}}} & \multicolumn{2}{c|}{\textbf{HuffPost}} \\ \cline{3-4} 
\multicolumn{1}{|c|}{} & \multicolumn{1}{c|}{} & \multicolumn{2}{c|}{5-way} \\ \cline{3-4} 
\multicolumn{1}{|c|}{} & \multicolumn{1}{c|}{} & \multicolumn{1}{c|}{1-shot} & 5-shot \\ \hline
LDS-PN w/o PT & None & \multicolumn{1}{c|}{66.2} & 74.5 \\ \hline
\multirow{5}{*}{LDS-PN w   PT} & This is a {[}MASK{]}   news: {[}sentence{]} & \multicolumn{1}{c|}{67.5} & \textbf{76.4} \\ \cline{2-4} 
 & A {[}MASK{]} news: {[}sentence{]} & \multicolumn{1}{c|}{67.9} & 76.3 \\ \cline{2-4} 
 & Topic: {[}MASK{]}. {[}sentence{]} & \multicolumn{1}{c|}{68.7} & 75.0 \\ \cline{2-4} 
 & Category: {[}MASK{]}. {[}sentence{]} & \multicolumn{1}{c|}{68.8} & 75.5 \\ \cline{2-4} 
 & {[}sentence{]} This topic is about {[}MASK{]}. & \multicolumn{1}{c|}{\textbf{68.9}} & 74.8 \\ \hline
\end{tabular}
\caption{Ablation study results of prompting (PT).}
\label{tb:template}
\end{table*}

\subsection{Step-up from PN to LDS-PN}
LDS-PN is a boosting version of PN with our LDS strategy. We make an ablation study step-up from PN to LDS-PN on the HuffPost and Clinc150 datasets. The results are shown in the following table, DS is our proposed Distance Scaling loss, LS is our Label-guided Scaler in the testing stage. As shown in Table.~\ref{tb:step}, PN

\subsection{The Effectiveness of Prompting}
\label{sec.pt}

We investigate the effectiveness of prompting by comparing LDS-PN with prompting and without prompting. We also investigate LDS-PN with prompting using different templates. Table~\ref{tb:template} shows the results under the 5-way 1-shot and 5-way 5-shot tasks on HuffPost dataset. It can be found that LDS-PN with prompting has around 2\% improvement compared to LDS-PN without prompting. And using different templates shows similar performances. Our key idea is that the LDS training phase treats the label representation as the distribution center of the classes so that when testing, we can further enhance the sample by using the known label names through the EM algorithm. Prompting serves to align the sample representations (sentence level) with the label representations (token level). Thus, the results of our approach do not depend primarily on prompting. Furthermore, we only used a basic manual design template, theoretically, some advanced methods of setting templates and extending label words could also be used in our LDS strategy, which may further enhance the effect of our approach.

\begin{figure*}[h!]
\centering
\subfigure[Prototypical Networks]{\includegraphics[width=0.32\textwidth]{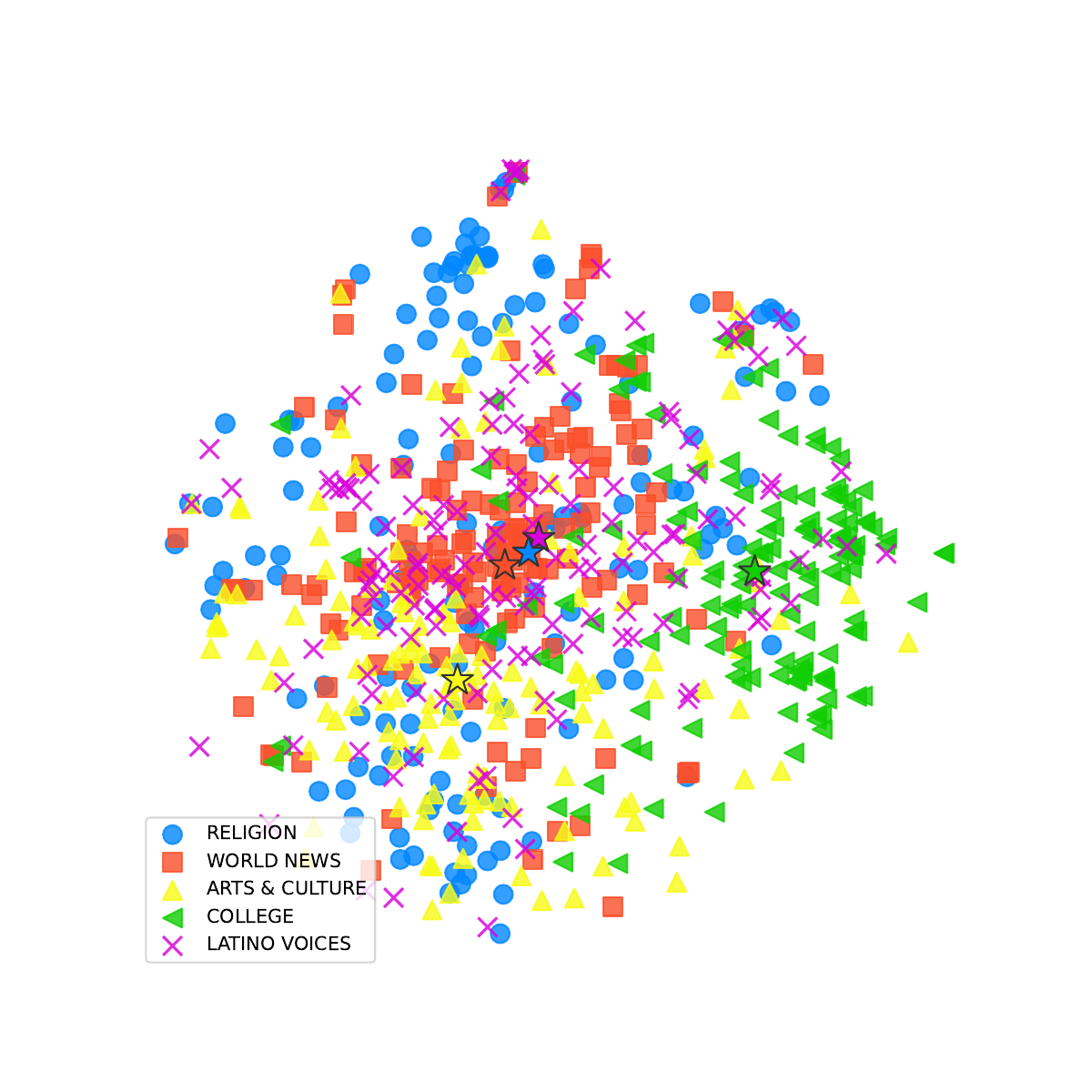}
\label{visa}}
\subfigure[ProtoVerb]{\includegraphics[width=0.32\textwidth]{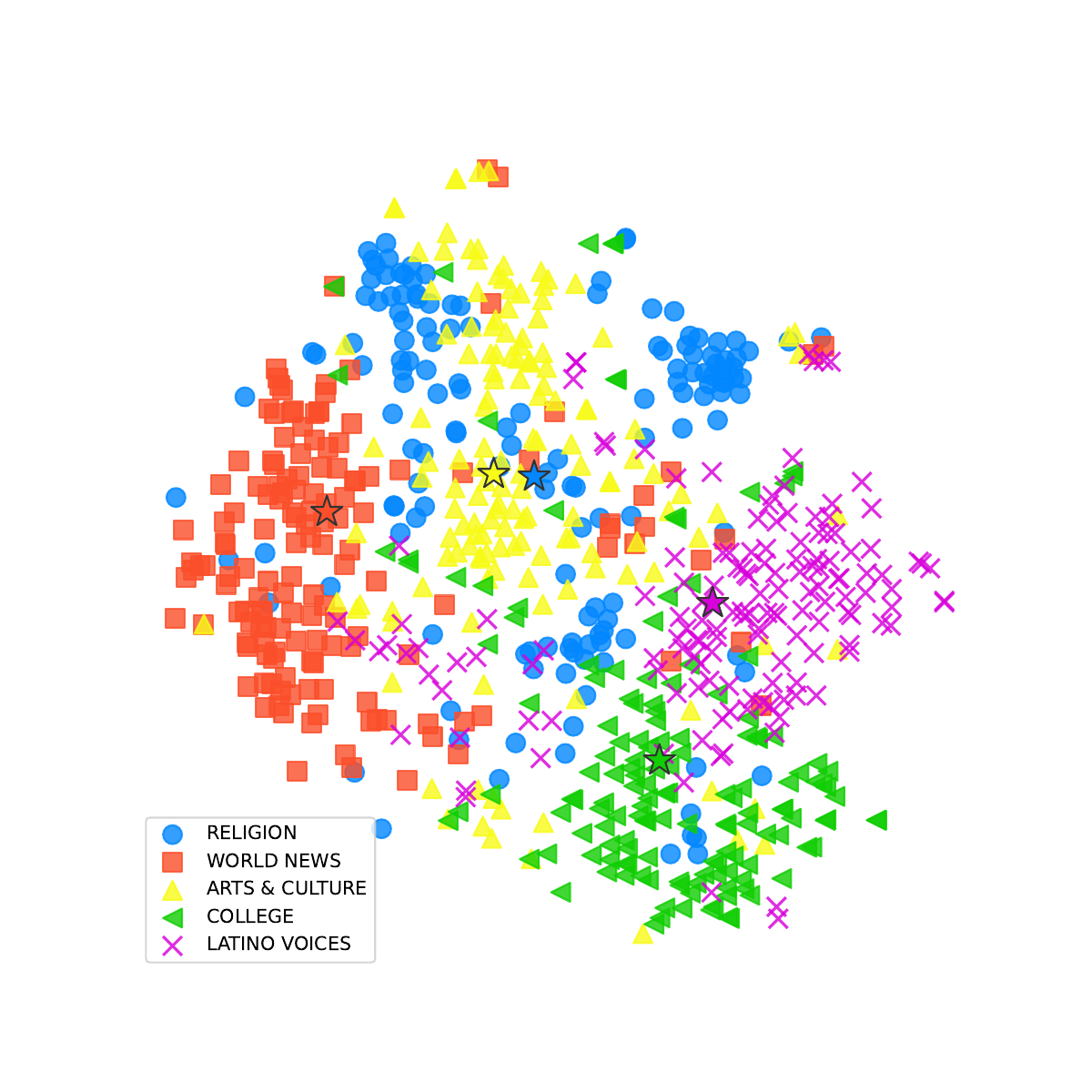}
\label{visb}}
\subfigure[LDS (ours)]{\includegraphics[width=0.32\textwidth]{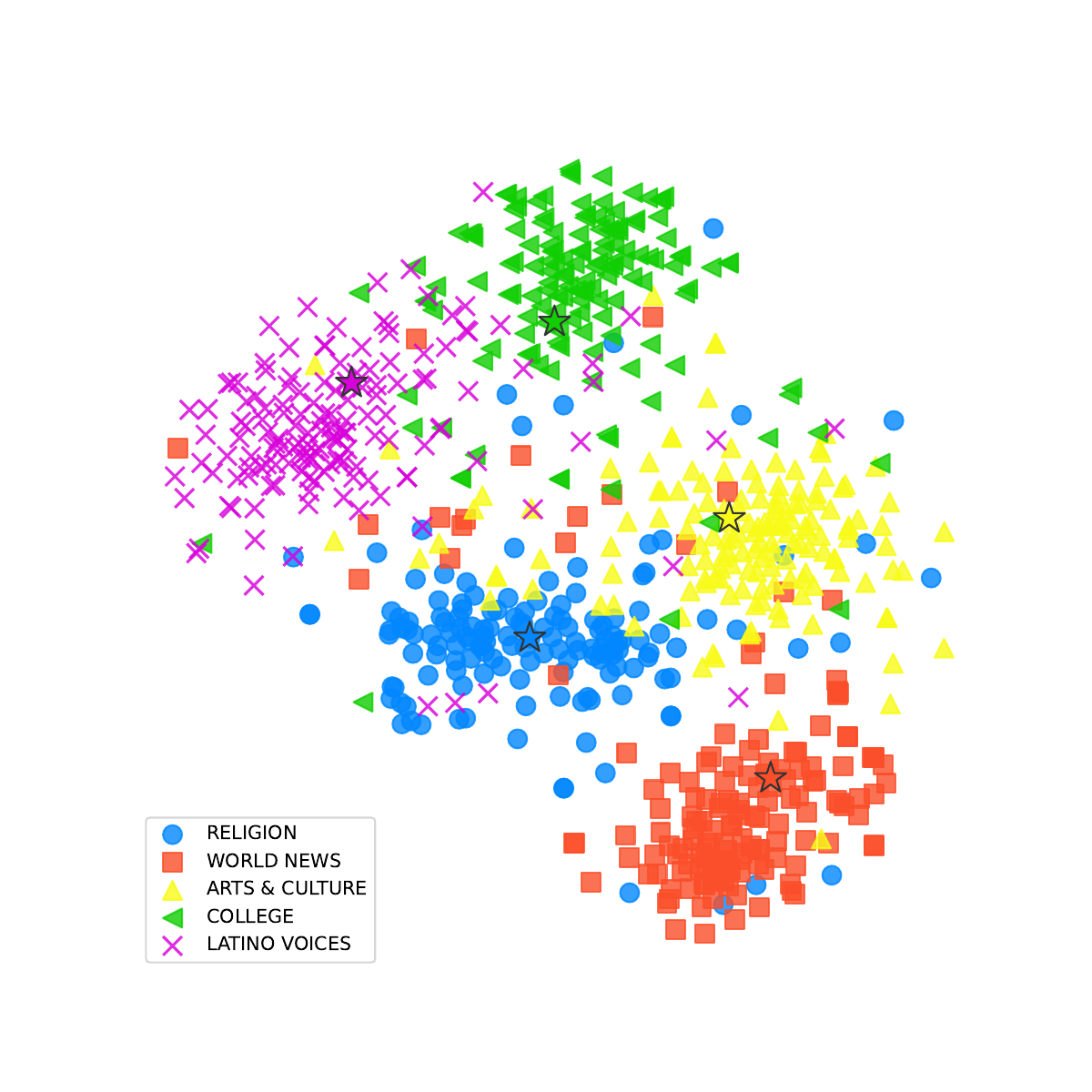}
\label{visc}}
\caption{Visualization of sample text representations sampled from five novel classes on HuffPost dataset. The input representations are given by (a) Prototypical Networks (b) ProtoVerb (a boosting version of PN by prompting and contrastive learning) and (c) LDS (ours). The pentagrams represent the class prototypes.}
\label{figvis2}
\end{figure*}

% \usepackage{multirow}
\begin{table*}[]
\centering
\begin{tabular}{|l|c|c|c|}
\hline
\textbf{Methods} &
  \multicolumn{1}{c|}{\textbf{Sentence}} &
  \multicolumn{1}{c|}{\textbf{Label names}} &
  \textbf{Cosine distance to the class center} \\ \hline
LDS &
  \multirow{2}{*}{\begin{tabular}[c]{@{}c@{}}Daily meditation : \\ a carefree heart\end{tabular}} &
  \multirow{2}{*}{RELIGION} & 0.53
   \\ \cline{1-1} \cline{4-4} 
LDS w/o LS &
   & 
   & 0.88
   \\ \hline
LDS & \multirow{2}{*}{\begin{tabular}[c]{@{}c@{}}12 money - saving ways  to \\ outsmart your grocery store\end{tabular}} & \multirow{2}{*}{TASTE} & 0.32  \\ \cline{1-1} \cline{4-4} 
LDS w/o LS & 
   & 
   & 0.69
   \\ \hline
LDS &
  \multirow{2}{*}{\begin{tabular}[c]{@{}c@{}}The top laptop deals \\ of black friday 2015\end{tabular}} &
  \multirow{2}{*}{TECH} & 0.59
   \\ \cline{1-1} \cline{4-4} 
LDS w/o LS & 
   & 
   & 0.70
   \\ \hline
LDS &
  \multirow{2}{*}{\begin{tabular}[c]{@{}c@{}}4 ways to get your\\  debt snowball rolling\end{tabular}} &
  \multirow{2}{*}{MONEY} & 0.66
   \\ \cline{1-1} \cline{4-4} 
LDS w/o LS & 
   & 
   & 0.68
   \\ \hline
LDS &
  \multirow{2}{*}{\begin{tabular}[c]{@{}c@{}}Art \& the tyranny\\  of the new\end{tabular}} &
  \multirow{2}{*}{ARTS} & 0.57
   \\ \cline{1-1} \cline{4-4} 
LDS w/o LS & 
   & 
   & 0.72
   \\ \hline
\end{tabular}
\caption{The case study result of a 5-way 1-shot task on HuffPost dataset.}
\label{tb:case}
\end{table*}

\section{Visualization}
To investigate models' ability to learn discriminative text representations, we visualize the sample representations and class prototypes produced by Prototypical Networks, ProtoVerb, and LDS using t-SNE~\cite{Maaten2008VisualizingDU}. We randomly select 5 novel classes on HuffPost dataset, where 50 samples as the support set and 100 samples as the query set per class. The results are shown in Fig.~\ref{figvis2}.  It can be observed that the text representation generated by ProtoVerb in Fig.~\ref{figvis2}(b) is more discriminative than that of the vanilla Prototypical Networks in Fig.~\ref{figvis2}(a). But the prototypes of the \textit{RELIGION} and \textit{ARTS \& CULTURE} classes are very close, making it hard to distinguish query samples between these two classes. As shown in Fig.~\ref{figvis2}(c), the text representations produced by our LDS are much more discriminative than the representations in Fig.~\ref{figvis2}(a) and Fig.~\ref{figvis2}(b). Moreover, the prototypes produced by LDS are easier to be distinguished and each prototype is closer to its class center, which can better represent its class.

\section{Case Study}
To demonstrate our main module Label-guided Scaler (LS) can mitigate the misclassification caused by randomly selected support samples, we randomly select 5 novel classes (150 samples per class) on HuffPost dataset, where 1 sample per class is selected as a support sample. As shown in Fig~\ref{casea} (LDS w/o LS), the support samples are far away from the class centers, but in Fig~\ref{caseb} (LDS), the support samples are pulled closer to the class centers. As shown in Table~\ref{tb:case}, by our LS module, support sample representations get a closer distance to the real class centers, especially when the support sample is near the boundary of class distribution (e.g., \textit{RELIGION} and \textit{TASTE} class).

\section{Results Analysis}
In addition to the main results of LDS-PN, we observe several interesting phenomena from Table \ref{tb:news} and Table \ref{tb:intent}, each of which is analyzed below.

\textbf{P1: The improvement of 1-shot is generally more than 5-shot.} In the 1-shot scenario, each class prototype is calculated through one support sample, because the selection of support sets is random, compared with the 5-shot scenario, it is more likely to result in inaccurate class prototypes, as shown in Fig.~\ref{fig1}(a). After our label-guided distance scaling (LDS), in the training stage, $\mathcal{L}_{LG}$ makes the sample representation as close as possible to the corresponding label representation, while away from other label representations. This strategy makes the label representation adhere to the distribution of the corresponding class. Therefore, in the test phase, the outlier sample representation can be pulled back into the distribution of the class by using the label representation through the EM algorithm. So each sample representation is closer to the center of the class by LDS, the better sample representations can be obtained and the better class prototypes can be calculated, as shown in Fig.~\ref{fig1}(b).

\textbf{P2: The 5-shot result on Amazon dataset is suboptimal.} Our method relies on the strong assumption that label representations obey the distribution of the class, which places higher demands on the quality of label names. To make a fair comparison, we only use the original labels in the datasets and do not further process the labels in the datasets. The Amazon dataset contains noisy reviews (1000 per class), and the label names may differ greatly from the actual centers. Although our LDS-PN does not exceed several strong baselines on the Amazon dataset in the 5-shot scenario, it is still best in the 1-shot scenario. And LDS-PN consistently outperforms vanilla PN.

\textbf{P3: The performance on the 10-way and 15-way tasks remains strong.} 
In the $N$-way $K$-shot setting, using a few amounts of labeled samples ($K$ samples per class), the more classes there are, the more difficult it is to classify. Some strong baselines such as ContrastNet, ProtoVerb, and TART, have shown strong performance in 5-way missions, but have not performed well in 10-way and 15-way tasks, especially the most recent method TART. In contrast, LDS-PN uses the two-stage LDS strategy of training and testing to generate sample representations that are closer to the class center, thus calculating better prototypes, achieving excellent performance in 10-way and 15-way tasks, and achieving 87.8\% accuracy in the 1-shot scenario, 90.3\% in the 5-shot scenario.

\end{document}